\documentclass[runningheads]{llncs}
\usepackage{graphicx}
\usepackage{amsmath,amssymb} 
\usepackage{color}
\usepackage[pagebackref=true,breaklinks=true,colorlinks,bookmarks=false]{hyperref}
\usepackage[width=122mm,left=12mm,paperwidth=146mm,height=193mm,top=12mm,paperheight=217mm]{geometry}

\graphicspath{{./figures/}}

\begin{document}
\pagestyle{headings}
\mainmatter

\title{Efficient Multi-Frequency Phase Unwrapping using Kernel Density Estimation} 

\titlerunning{Efficient Multi-Frequency Phase Unwrapping using KDE}

\authorrunning{Felix J{\"a}remo Lawin \and Per-Erik Forss{\'e}n \and Hannes Ovr{\'e}n}

\author{Felix J{\"a}remo Lawin \and Per-Erik Forss{\'e}n \and Hannes Ovr{\'e}n}

\institute{Computer Vision Laboratory, Link{\"o}ping University, Sweden\\ \texttt{\{felix.jaremo-lawin, per-erik.forssen, hannes.ovren\}@liu.se}}

\maketitle

\begin{abstract}
	In this paper we introduce an efficient method to unwrap multi-frequency phase estimates for time-of-flight ranging. The algorithm generates multiple depth hypotheses and uses a spatial kernel density estimate (KDE) to rank them. The confidence produced by the KDE is also an effective means to detect outliers. 
	We also introduce a new closed-form expression for phase noise
	prediction, that better fits real data. 
	The method is applied to depth
	decoding for the Kinect v2 sensor, and compared to the {\it
		Microsoft Kinect SDK} and to the open source driver {\it
		libfreenect2}. 
	The intended Kinect v2 use case is scenes with less than $8$m range, and for such cases we observe consistent improvements, while maintaining real-time performance.
	When extending the depth range to the maximal value of $18.75$m, we get about $52\%$ more valid measurements than {\it libfreenect2}.
	The effect is that the sensor can now be used in large depth scenes, where it was previously not a good choice. Code and supplementary material are available at \url{http://www.cvl.isy.liu.se/research/datasets/kinect2-dataset/}.
	
\end{abstract}

\section{Introduction}

Multi-frequency time-of-flight is a way to accurately estimate
distance, that was originally invented for Doppler RADAR \cite{trunk93}. More recently it has also found an application in RGB-D sensors\footnote{RGB-D  sensors output both colour (RGB) and depth (D) images.} that use time-of-flight ranging, such as the {\it Microsoft Kinect v2} \cite{sell14}.

Depth from time-of-flight requires very accurate time-of-arrival estimation.
Amplitude modulation improves accuracy, by measuring phase
shifts between the received and emitted signals, instead of
time-of-arrival. However, a disadvantage with amplitude modulation is
that it introduces a periodic depth ambiguity. By using multiple
modulation frequencies in parallel, the ambiguity can be resolved in
most cases, and the useful range can thus be extended.

\begin{figure}[!t]
	\begin{center}
		\scalebox{-1}[1]{\includegraphics[height=0.23\columnwidth] {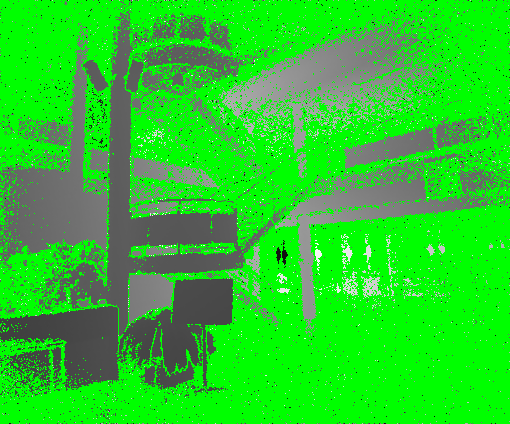}}
		\scalebox{-1}[1]{\includegraphics[height=0.23\columnwidth] {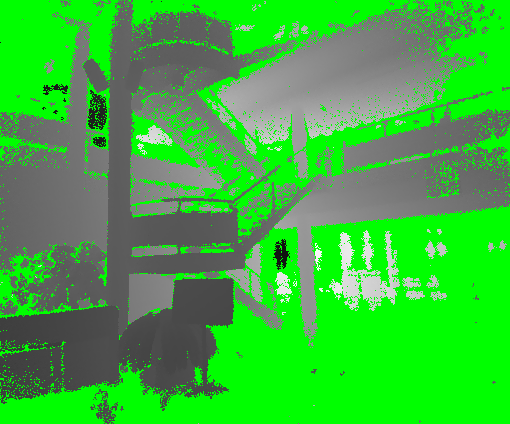}}
		\scalebox{-1}[1]{\includegraphics[height=0.23\columnwidth] {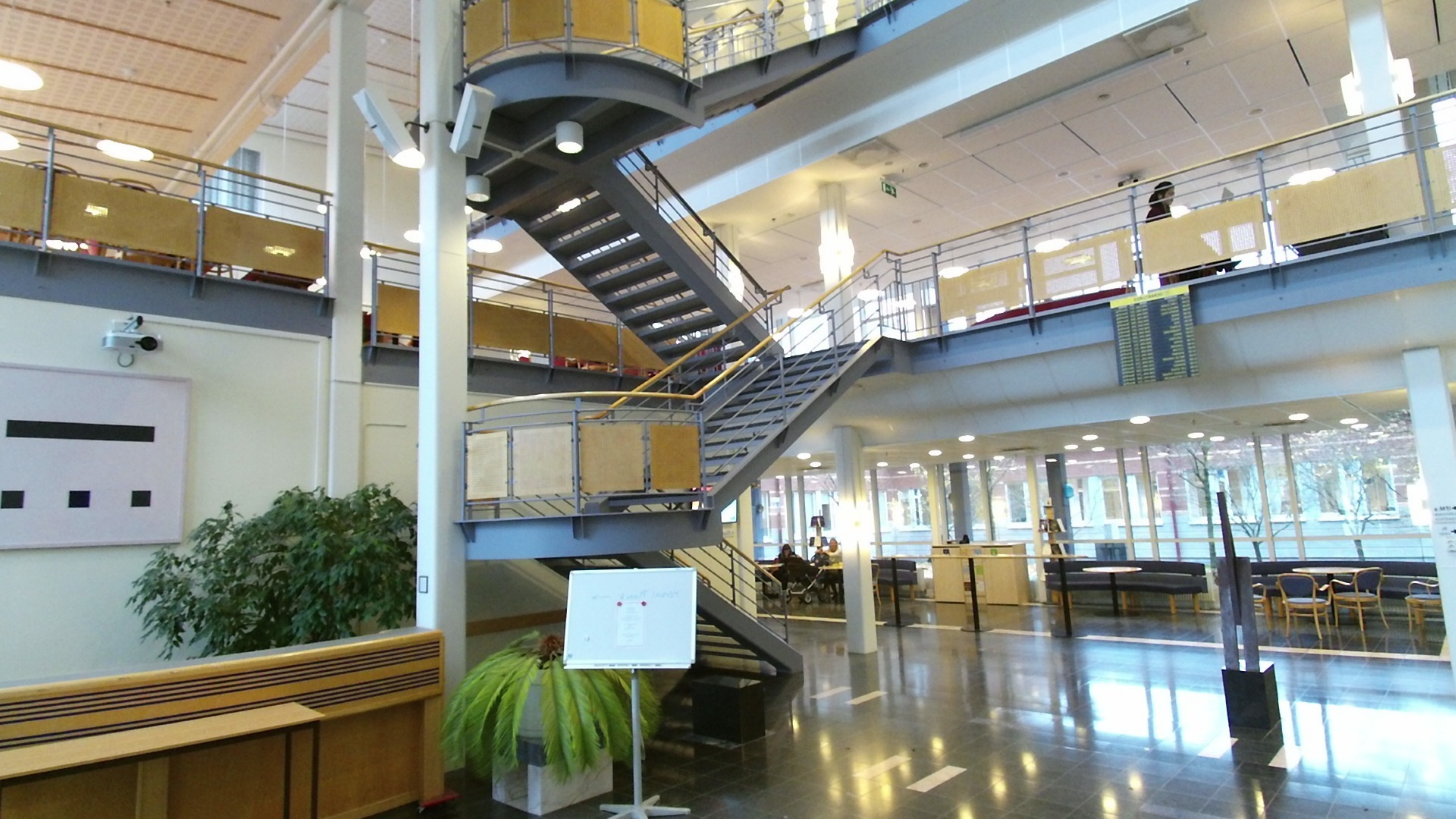}}
		\caption{Single frame output on scene with greater than $18.75$m depth range. Left:
			{\it libfreenect2}, Center: proposed method. Right: corresponding RGB image. Pixels suppressed by outlier rejection are shown in green. The proposed method has more valid depth points than {\it libfreenect2} resulting in a denser and more well defined depth scene. While the suppressed areas are clean from outliers for the proposed method, the {\it libfreenect2} image is covered in salt and pepper noise. 
		}
		\label{fig:mesh}
	\end{center}
\end{figure}

We introduce an efficient method to unwrap multi-frequency phase estimates for time-of-flight ranging. The
algorithm uses {\it kernel density estimation} (KDE) in a spatial
neighbourhood to rank different depth hypotheses. The KDE also
doubles as a confidence measure
which can be used to detect and
suppress bad pixels. We apply our method to depth decoding for the
Kinect v2 sensor. For large depth scenes we see a significant increase
in coverage of about $52\%$ more valid pixels compared to {\it
	libfreenect2}. See figure \ref{fig:mesh} for a qualitative
comparison. 
For 3D modelling with Kinect fusion \cite{newcombe11}, this results in fewer outlier points and more complete scene details.
While the method is designed with the Kinect v2 in mind, it is also applicable to multi-frequency ranging techniques in general.

\subsection{Related Work}

The classic solution to the multi-frequency phase unwrapping problem, is to use the Chinese reminder theorem (CRT). This method is fast, but implicitly assumes noise free data, and in \cite{trunk93} it is demonstrated that by instead generating multiple unwrappings for each frequency, and then performing clustering along the range axis, better robustness to noise is achieved. However, due to its simplicity, CRT is still advocated, e.g.~in \cite{jongenelen11,wang2009}, and is also used in the Kinect v2 drivers.

Simultaneous unwrapping of multiple phases with different frequencies
is a problem that also occurs in fringe pattern projection techniques
\cite{gorthi10,wang2009}. The algorithms are not fully equivalent though,
as the phase is estimated by different means, and the relationship
between phase and depth is different. 

Another way to unwrap the time-of-flight phase shift is to use surface reflectivity constraints. As the amplitude associated with each phase measurement is a function of object distance and surface reflectivity, a popular approach in the literature is to assume locally constant reflectivity. Under this assumption, the depth can be unwrapped using e.g.~a Markov Random Field (MRF) formulation with a data term and a reflectivity smoothness term. In \cite{hansard13},
many different such unwrapping methods are discussed.  A recent extension of this is \cite{crabb15}, where distance, surface albedo and also the local surface normal are used to predict the reflectance.

The multi-frequency and reflectivity approaches are combined in \cite{droeschel10} where a MRF with both reflectivity, and dual frequency data terms are used.

Detection of {\it multipath interference} (i.e.~measurement problems due to light reflected from several different world locations reaching the same pixel) is studied in \cite{kirmani13}. If four or more frequencies are used, pixels with multipath effects can be detected and suppressed. Recently in \cite{feigin15}, a multipath detection algorithm based on blind source separation was applied to the Kinect v2. This required the firmware of the Kinect v2 to be modified to emit and receive at 5 frequencies instead of the default 3.
As firmware modification currently requires reverse engineering of the
transmission protocol we have not pursued this line of research.

In \cite{mei13} a simulator for ToF measurements is developed and used
to evaulate performance of a MRF that does simultaneous unwrapping and
denoising using a wavelet basis. The performance on real data is however not shown.

Noise on the phase measurements is analyzed in \cite{frank09} and it
is suggested that the variance of phase is predicted by sensor
variance divided by the phase amplitude squared. In this paper we
derive a new model for phase noise that fits better with real
data and utilize it as a measure of confidence for the measurements. In \cite{mei13} a Gaussian mixture model for sensor noise is
also derived, but its efficacy is never validated on real sensor data.

\subsection{Structure}

The paper is organized as follows: In section \ref{sec:depth_decoding}
we describe how multi-frequency time-of-flight measurements are used to sense depth. In section \ref{sec:kde} we describe how we extend
this by generating multiple hypotheses and selecting one based on 
kernel density estimation. We give additional implementation details and compare our method to other
approaches in section \ref{sec:experiments}. The paper concludes with a
discussion and outlook in section \ref{sec:conclusions}.

\section{Depth Decoding}
\label{sec:depth_decoding}

In time-of-flight sensors, an amplitude modulated light signal is
emitted to be reflected on objects in the environment. The reflected
signal is then captured in the pixel array of the sensor, where it is
correlated with the reference signal driving the light emitter. On the
Kinect v2 this is achieved on the camera chip by using quantum
efficiency modulation and integration\cite{patent740,bamji15}
resulting in a voltage value $v_k$. In the general case $N$ different
reference signals are used, each phase shifted $\frac{2\pi}{N}$
radians from the others \cite{frank09}. Often $N=4$ is used
\cite{mei13}, but in the Kinect v2 we have $N=3$. The voltage values are used to calculate the phase shift between the emitted and the received signals using the complex phase

\begin{equation}
{\bf z}=\frac{2}{N}\sum_{k=0}^{N-1} v_ke^{-{\bf i}(p_o+2\pi k/N)}\,,
\end{equation}
where $p_o$ is a common phase offset. This expression is derived using
least squares \cite{frank09}, and the actual phase shift and its
corresponding amplitude are obtained as
\begin{equation}
\phi=\textrm{arg}\;{\bf z}\quad\textrm{and}\quad a=|{\bf z}|\,.
\label{eq:phase_amplitude}
\end{equation}
The amplitude is proportional to the reflected signal strength, and
increases when the voltage values make consistent contributions to
${\bf z}$. It is thus useful as a measure of confidence in the decoded
phase.

From the phase shift $\phi$ in \eqref{eq:phase_amplitude} the
time-of-flight distance can be calculated as
\begin{equation}
d=\frac{c\phi}{4\pi f_m}\,,
\label{eq:phase_depth}
\end{equation}
where $c$ is the speed of light, and $f_m$ is the used modulation
frequency (see e.g.~\cite{sell14}).  This relationship holds both in
multi-frequency RADAR \cite{trunk93} and RGB-D time-of-flight. In
fringe projection profilometry \cite{gorthi10}, phase and amplitude
values are also obtained for each frequency of the fringe pattern,
resulting in a very similar problem. However, the relationship between
phase and depth is different in this case.

The phase shift obtained from \eqref{eq:phase_amplitude} is the true
phase shift $\tilde{\phi}$ modulo $2\pi$. Thus $\phi$ is ambiguous in an environment
where $d$ can be larger than $c/(2f_m)$. Finding the correct period,
i.e.~$n$ in the expression
\begin{equation}
\tilde{\phi}=\phi_\textrm{wrapped}+2\pi n\,,\quad n\in\mathbb{N}\,,
\label{eq:unwrapping}
\end{equation}
is called {\it phase unwrapping}.
To reduce measurement noise, and increase the range in which $\phi$ is unambiguous, one can combine the phase measurements from multiple modulated signals with different frequencies.

\begin{figure}[t!]
	\begin{center}
		\includegraphics[width=0.6\columnwidth] {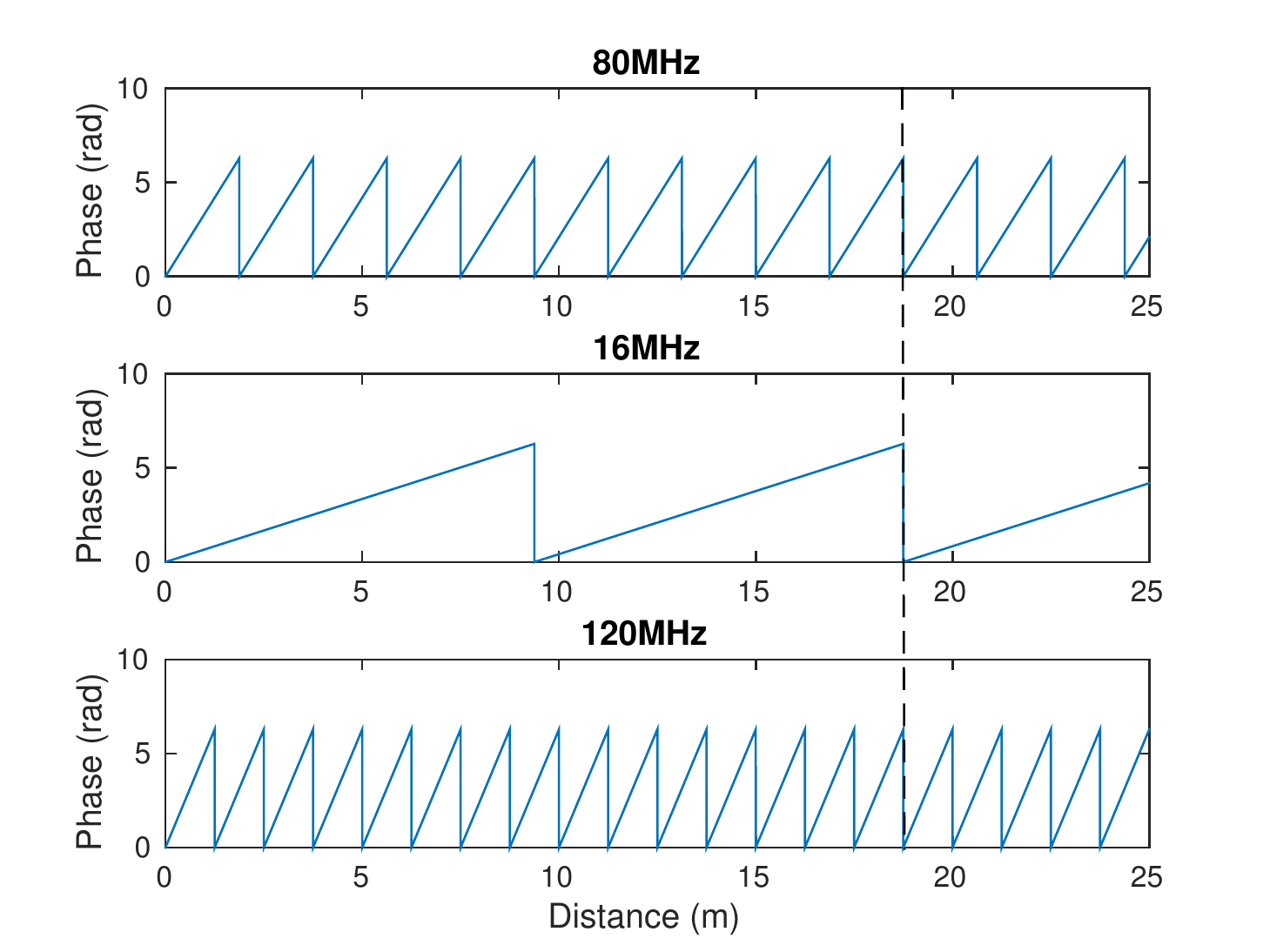} 

		\caption{Wrapped phases for Kinect v2, in the range 0 to 25
			meters. Top to bottom: $\phi_0$, $\phi_1$,
			$\phi_2$. The dashed line at 18.75 meters indicates
			the common wrap-around point for all three
			phases. Just before this line we have $n_0=9$, $n_1=1$,
			and $n_2=14$.}
		\label{fig:phasetodist}
	\end{center}
	
\end{figure}

Figure \ref{fig:phasetodist} shows the phase to distance relation for
the three amplitude-modul-ated signals, with frequencies 16, 80 and 120
MHz, which is the setup used in the Kinect v2 \cite{sell14}. For each of the three
frequencies, three phase shifts are used to calculate a phase according to
\eqref{eq:phase_amplitude}, and thus a total of nine measurements are
used in each depth calculation.
In the figure, we see that if the phase shifts are combined, a common wrap-around occurs at 18.75 meters. This is thus the maximum range in which the Kinect v2 can operate without depth ambiguity.

As a final step, the phase shifts from the different modulation frequencies
are combined using a phase unwrapping procedure and a weighted
average. 

It is of critical importance that the phase is correctly unwrapped, as
choosing the wrong period will result in large depth errors. This is
the topic of the following sub-sections.

\subsection{Phase unwrapping}
Consider phase measurements of $M$ amplitude modulated signals with different modulation frequencies. From \eqref{eq:phase_depth} we get the following relations:

\begin{align}
& d = \frac{c \left (\phi_0+2 \pi n_0 \right)}{4 \pi f_0} = \frac{c \left (\phi_1+2 \pi n_1 \right)}{4 \pi f_1} = \dots = \frac{c \left (\phi_{M-1}+2 \pi n_{M-1} \right)}{4 \pi f_{M-1}} \iff\\
& \frac{k_0}{2 \pi } \phi_0 + k_0 n_0 = \frac{k_1}{2 \pi } \phi_1 + k_1 n_1 = \ldots =\frac{k_{M-1}}{2 \pi } \phi_{M-1} + k_{M-1} n_{M-1} 
\label{eq:general_unwrapped_phases_to_dist}
\end{align}
where $\{ k_m \}^{M-1}_{m=0}$ are the least common multiples for $\{f_m\}^{M-1}_{m=0}$ divided by the respective frequency and $\{ n_m \}^{M-1}_{m=0}$ are the set of sought unwrapping coefficients. Now \eqref{eq:general_unwrapped_phases_to_dist} can be simplified to a set of constraints on pairs of unwrapping coefficients $\left(n_i, n_j \right)$:

\begin{equation}
k_i n_i -k_j n_j = \frac{k_{j}}{2 \pi } \phi_{j} - \frac{k_{i}}{2\pi } \phi_{i}, \forall i,j \in [0,M-1] \textrm{ and } i > j.
\label{eq:sys_general}
\end{equation}

In total there are $M(M-1)/2$ such equations. As the system is redundant, the correct unwrapping cannot be obtained
by e.g.~Gaussian elimination and in practice the equations are
unlikely to hold due to measurement noise. The constraints can however be
used to define a likelihood for a specific unwrapping.

\subsection{CRT based unwrapping}
\label{sec:greedy}
The ambiguity of the phase measurements can be resolved by applying a
variant of the Chinese reminder theorem (CRT) \cite{jongenelen11,wang2009} to one equation at a time in \eqref{eq:sys_general}:

\begin{align}
\label{eq:CRT}
n_i &= k_i \cdot \textrm{round}\left(\frac{k_j \phi_j-k_i\phi_i}{k_i 2 \pi }\right)\\
\tilde{\phi}_i &= \phi_i + 2 \pi n_i 
\end{align}

In the case of more than two frequencies the unwrapped phase $\tilde{\phi}_i$ could be used in \eqref{eq:CRT} for the next equation in \eqref{eq:sys_general} to unwrap the next phase. This is suggested and described in \cite{wang2009} and is also used in {\it libfreenect2}. In the end when all equations have been used, the full unambiguous range of the combined phase measurements has been unwrapped. The CRT method is fast but sensitive to noise as it unwraps each of the phase measurements in sequence. The consequence of this is that an error made early on will be propagated. 

\subsection{Phase fusion}
\label{sec:phase_fusion}
The unwrapped phase measurements are combined by using a weighted average:
\begin{equation}
t^\ast = \sum_{m=0}^{M-1} \frac{k_m \tilde{\phi}_m}{(k_m\sigma_{\phi_m})^2}\Big/\left(\sum_{m=0}^{M-1}\frac{1}{(k_m\sigma_{\phi_m})^2}\right)\,,
\label{eq:fused_phase}
\end{equation}

where $\sigma_{\phi_m}$ is the standard deviation of the noise in $\phi_m$. The {\it pseudo distance} estimate $t^\ast$ is later  converted to a depth (i.e.~distance in the forward direction), using
the intrinsic camera parameters.

\section{Kernel Density based Unwrapping}
\label{sec:kde}

In this paper, we propose a new method for multi-frequency phase unwrapping.
The method considers several fused pseudo distances
$t^\ast$ (see \eqref{eq:fused_phase}) for each pixel location
${\bf x}$, and select the one with the highest kernel density value
\cite{murphy12}. Each such hypothesis $t^i({\bf x})$ is a function of
the unwrapping coefficients ${\bf
	n}=(n_0,\ldots, n_{M-1})$. The kernel
density for a particular hypothesis $t^i({\bf x})$ is a weighted sum
of all considered hypotheses in the spatial neighbourhood:
\begin{equation}
p(t^i({\bf x}))=\frac{\sum_{j\in\mathcal{I},k\in \mathcal{N}({\bf x})} w_{jk}K(t^i({\bf
		x})-t^j({\bf x}_k))}{\sum_{j\in\mathcal{I},k\in \mathcal{N}({\bf x})}
	w_{jk}}\,.
\label{eq:kde}
\end{equation}
Here $K(\cdot)$ is the kernel, and $w_{jk}$ is a sample weight. The
sets of samples to consider are defined by the hypothesis indices $\mathcal{I}$
(e.g.~$\mathcal{I}=\left\{1,2\right\}$ if we have two hypotheses in each pixel),
and by the set of all spatial neighbours $\mathcal{N}({\bf x})=\left\{k:\|{\bf x}_k-{\bf x}\|_1<r\right\}$ where $r$ is a
square truncation radius. The hypothesis weight $w_{ik}$ is defined as
\begin{equation}
w_{ik}=g({\bf x}-{\bf x}_k,\sigma)p(t^i({\bf x}_k)|{\bf n}_i({\bf x}_k))p(t^i({\bf x}_k)|{\bf a}_i({\bf x}_k))\,.
\label{eq:weights}
\end{equation}
The three factors in $w_{ik}$ are:
\begin{itemize}
	\item the {\it spatial weight} $g({\bf x}-{\bf x}_k,\sigma)$, which is a
	Gaussian that downweights neighbours far from the considered pixel location
	${\bf x}$.
	\item the {\it unwrapping likelihood} $p(t^i({\bf x})|{\bf
		n}_i({\bf x}))$, that depends on the consistency of the pseudo-distance estimate \eqref{eq:fused_phase} given the unwrapping vector \\${\bf n}=(n_0,\ldots, n_{M-1})$.
	\item the {\it phase likelihood} $p(t^i({\bf x})|{\bf a}_i({\bf
		x}))$, where ${\bf a}_i=(a_0,\ldots, a_{M-1})$, are the
	amplitudes from \eqref{eq:phase_amplitude}. It defines
	the accuracy of the phase before unwrapping.
\end{itemize}

The kernel in \eqref{eq:kde} is defined as:

\begin{equation}
K(x) = e^{-x^2/2 h^2}\,,
\label{eq:the_kernel}
\end{equation}
where $h$ is the kernel scale.

In the following sub-sections we will describe the three weight terms in
more detail. For simplicity of notation, we will drop the pixel
coordinate argument ${\bf x}$, and e.g.~write $p(t^\ast)$ instead of
$p(t^\ast({\bf x}))$.

\subsection{Unwrapping likelihood}
\label{sec:unwrapping_likelihood}

Due to measurement noise, the constraints in \eqref{eq:sys_general} are never perfectly satisfied. We thus subtract the left-hand side from the
right-hand side of these equations to form residuals $\epsilon_k$,
one for each of the $M(M-1)/2$ constraints. These are then used to define
a cost for a given unwrapping vector ${\bf
	n}=(n_0,\ldots, n_{M-1})$:
\begin{equation}
J({\bf n}) = \sum_{k=1}^{M(M-1)/2}\epsilon_k^2/\sigma_{\epsilon_k}^2\,.
\label{eq:knm_cost}
\end{equation}

This cost function corresponds to the following {\it unwrapping likelihood}:
\begin{equation}
p(t^\ast|{\bf n})\propto e^{-J({\bf n})/(2s_1^2)}\,,
\label{eq:unwrapping_likelihood}
\end{equation}
where $t^\ast$ is the fusion of the three unwrapped pseudo-distances,
see \eqref{eq:fused_phase}, and $s_1$ is a scaling factor to be
determined. For  normally distributed residuals, and the Kinect v2
case of $M=3$, the constraints in \eqref{eq:sys_general} imply:
\begin{align}
\sigma_{\epsilon_1}^2 &= \left(\frac{k_1\sigma_{\phi_1}}{2\pi}\right)^2+\left(\frac{k_0\sigma_{\phi_0}}{2\pi}\right)^2 \label{eq:weights1}\\
\sigma_{\epsilon_2}^2 &= \left(\frac{k_2\sigma_{\phi_2}}{2\pi}\right)^2+\left(\frac{k_0\sigma_{\phi_0}}{2\pi}\right)^2\\
\sigma_{\epsilon_3}^2 &= \left(\frac{k_2\sigma_{\phi_2}}{2\pi}\right)^2+\left(\frac{k_1\sigma_{\phi_1}}{2\pi}\right)^2\,.
\label{eq:weights3}
\end{align}
This gives us the weights in \eqref{eq:knm_cost}. 
The values of $\sigma_{\phi_m}$ could be predicted from the phase
amplitude $a_m$ (more on this later), but they tend to deviate around
a fixed ratio, and we have observed better robustness of
\eqref{eq:unwrapping_likelihood} if the ratio is always
fixed. We assume that the phase variances is equal for all modulation frequencies. This assumption gives us their relative magnitudes, but not their
absolute values, which motivates the introduction of the parameter
$s_1$ in \eqref{eq:unwrapping_likelihood}.

\subsection{Multiple hypotheses}
\label{sec:multiple_hypotheses}
In contrast to the CRT approach to unwrapping, see section
\ref{sec:greedy}, we will consider all meaningful unwrapping vectors ${\bf n}=(n_0,\ldots, n_{M-1})$ within the unambiguous range. A particular depth value corresponds to a unique unwrapping vector, but with the introduction of noise, neigbouring unwrappings need to be considered at wrap around points. For example, looking at the Kinect v2 case shown in figure \ref{fig:phasetodist}, if $n_0 = n_1 = 0$,  $n_2$ should either be 0 or 1. In total $30$ different hypotheses for $(n_0,n_1,n_2)$ are constructed in this way. These can then be ranked by
\eqref{eq:unwrapping_likelihood}.

Compared with the CRT approach, that only considers
one hypothesis, the above approach is more expensive. On the other
hand, the true maximum of \eqref{eq:unwrapping_likelihood} is guaranteed to be checked.

In the low noise case, we can expect the hypothesis with the largest
likelihood according to \eqref{eq:unwrapping_likelihood} to be the
correct one. This is however not necessarily the case in
general. Therefore a subset $\mathcal{I}$ of hypotheses with high
likelihoods are saved for further consideration, by evaluating the
full kernel density \eqref{eq:kde}.

\subsection{Phase likelihood}
\label{sec:phase_likelihood}
The amplitude, $a$, produced by \eqref{eq:phase_amplitude} can be used to
accurately propagate a noise estimate on the voltage values to noise
in the phase estimate. In \cite{frank09} this relationship is analysed
and an expression is derived that can only be computed numerically.
For practical use, \cite{frank09} instead propose $\sigma^2_\phi = 0.5(\sigma_v/a)^2$ as
approximate propagation formula (for $N=4$).

For constant but unknown noise variance on the voltage values $\sigma^2_v$, the phase noise can be predicted from the amplitude, as:
\begin{equation}
\sigma_\phi = \gamma/a\,,
\label{eq:sigma_frank}
\end{equation}
where $\gamma$ is a parameter to be determined. While propagation of
noise from voltage values to the complex phase vector ${\bf z}$ is linear, the final phase extraction is
not, and we will now derive a more accurate approximation using sigma-point 
propagation \cite{uhlmann95}. Geometrically, phase
extraction from the phase vector \eqref{eq:phase_amplitude} is a
projection onto a circle, and thus the noise propagation is also a
projection of the noise distribution $p({\bf z})$ onto the circle, see figure
\ref{fig:propagation_models} (a). $p({\bf z})$ is centered around the true
amplitude $a$, and sigma-point candidates are located on a circle with
radius $\sigma_z$. By finding the points where the circle tangents
pass through the origin, we get an accurate projection of the noise distribution.

The points of tangency can be found using the pole-polar
relationship \cite{hartley03}. 
For points $(x,y)$ and $(x,-y)$ we get the expressions:

\begin{equation}
x =(a^2-\sigma_z^2)/a\quad\textrm{and}\quad y =\frac{\sigma_z}{a}\sqrt{a^2-\sigma_z^2}\,.
\end{equation}
From these expressions, the phase noise can be predicted as:
\begin{equation}
\hat{\sigma}_\phi=\tan^{-1}(y/x)=\tan^{-1}(\sqrt{1/((a/\sigma_z)^2-1)})\,,
\label{eq:sigma_raw}
\end{equation}
where $\sigma_z$ is a model parameter to be determined. Values of
$a<\sigma_z$ invalidate the geometric model in figure
\ref{fig:propagation_models} (a), and for these we use \eqref{eq:sigma_frank} with $\gamma = \sigma_z \pi/2$.

In {\it libfreenect2}, a bilateral filter is applied to the ${\bf z}$
vectors. The noise attenuation this results in is amplitude dependent,
but it can be accurately modelled as a quadratic polynomial on $a$.
\begin{equation}
\hat{\sigma}_{\phi,\text{bilateral}}=\tan^{-1}(y/x)=\tan^{-1}(\sqrt{1/((\gamma_0+a\gamma_1+a^2\gamma_2)^2-1)})\,,
\label{eq:sigma_bilateral}
\end{equation}

We use the predicted phase noise to define a {\it phase likelihood}:
\begin{equation}
p(t^\ast|{\bf a})=\prod^{M-1}_{m=0}p(t^\ast|a_m)\,,\text{
	where } p(t^\ast|a_m)\propto e^{-0.5\hat{\sigma}^2_{\phi_m}/s^2_2}\,.
\label{eq:phase_likelihood}
\end{equation}
where $s_2$ is a parameter to be tuned. The phase likelihood encodes the accuracy of the phases {\it before} unwrapping.

\subsection{Hypothesis selection}
\label{sec:hypothesis_selection}
In each spatial position ${\bf x}$, we rank the considered hypotheses
$t^i$, using the KDE (kernel density estimate) defined in \eqref{eq:kde}.
The final hypothesis selection is then made as:
\begin{equation}
i^\ast=\arg\max_{i\in\mathcal{I}} p(t^i)\,.
\label{eq:selection}
\end{equation}
For the selected hypothesis, $p(t^{i^\ast})$ is also useful as a confidence measure
that can be thresholded to suppress the output in problematic pixels.
However, if the spatial support is small, e.g.~$3\times 3$, the weighted KDE
occasionally encounters sample depletion problems (only very bad
samples in a neighbourhood). This can be corrected by regularizing the
confidence computation according to:
\begin{equation}
\text{conf}(t^i)=\frac{\sum_k w_kK(t^i-t^k)}{\max(p_\text{min},\sum_k
	w_k)}\approx p(t^i)\,,
\label{eq:confidence}
\end{equation}
where $p_\text{min}$ is a small value, e.g.~$0.5$.

\subsection{Spatial selection versus smoothing}
\label{sec:spatial_kde}
The proposed KDE approach, see \eqref{eq:kde}, selects the best phase unwrapping
by considering the distribution of hypotheses in the spatial
neighbourhood of a pixel. Note that the spatial neighbourhood is only
used to {\it select}  among different hypotheses. This is different
from a spatial smoothing, as is commonly used in e.g.~depth from
disparity \cite{szeliski10}. A connection to kernel based smoothing approaches, such as
channel smoothing \cite{forssen04,ffs05} can be made by considering
the limit where the number of hypotheses is the continuous set of
$t$-values in the depth range of the sensor. The discrete selection in
\eqref{eq:selection} will then correspond to decoding of the highest
peak of the PDF, and thus to channel smoothing. In the experiments we
will however use just $|\mathcal{I}|=2$, or $3$ hypotheses per pixel,
which is far from this limit.
After selection, the noise on each pixel is still uncorrelated from the
noise of its neighbours, and each pixel can thus still be considered
an independent measurement. This is beneficial when fusing
data in a later step, using e.g.~Kinect Fusion \cite{newcombe11}.

\section{Experiments}
\label{sec:experiments}
We apply the method to depth decoding for the Kinect v2 sensor, and compare it to the {\it Microsoft Kinect SDK}\footnote{Version
	2.0.1409} in the following denoted {\it Microsoft} and to the open source driver {\it libfreenect2}.
A first visual result is shown in figure \ref{fig:mesh}. As can be seen in the
figure, the proposed method has a better coverage in the depth images
than {\it libfreenect2}. Another clear distinction between the
methods is that  {\it libfreenect2} produces salt and pepper noise all
over the image. See also \cite{jaremo-lawin16a} for more examples, and
corresponding RGB frames.

\subsection{Implementation}
\label{sec:implementation}
The algorithm was implemented by modifying the {\it
	libfreenect2}\footnote{As in $opencl\_depth\_packet\_processor.cl$ Feb 18 2016 commit: 1d06d2db04a9 } code for depth calculations using {\it OpenCL}
\cite{opencl} for GPU acceleration. When running the proposed pipeline with $|\mathcal{I}|=2$ on a {\it Nvidia GeForce GTX 760} GPU, the frame rate for the depth calculations is above 30 fps for spatial supports up to $17\times
17$. For e.g.~a $3\times 3$ support our method operates at 200 fps,  which is marginally slower than {\it
	libfreenect2} (which also operates in a $3\times 3$ neighbourhood) at $245$ fps. The current implementation is
however designed for ease of testing, and further speed optimization is
be possible.

\begin{figure}[!t]
	\begin{tabular}{@{}c@{\;}c@{\;}c@{}}
		{\bf library}(tuning) & {\bf kitchen}(test) & {\bf lecture}(test) \\
		\includegraphics[width=0.325\columnwidth]{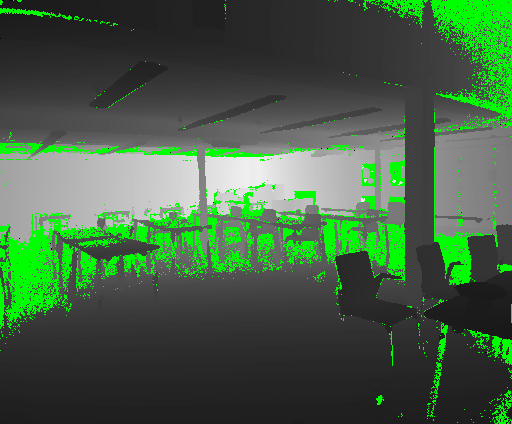} &
		\includegraphics[width=0.325\columnwidth]{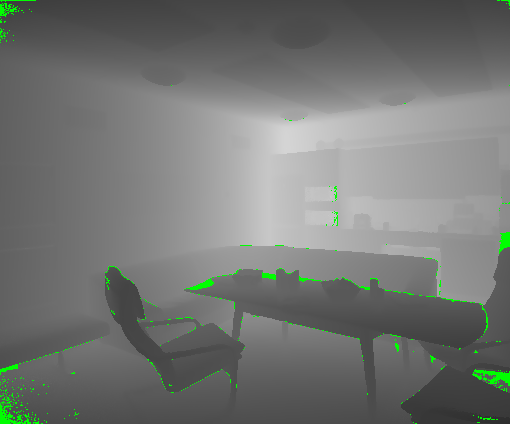} &
		\scalebox{-1}[1]{\includegraphics[width=0.325\columnwidth]{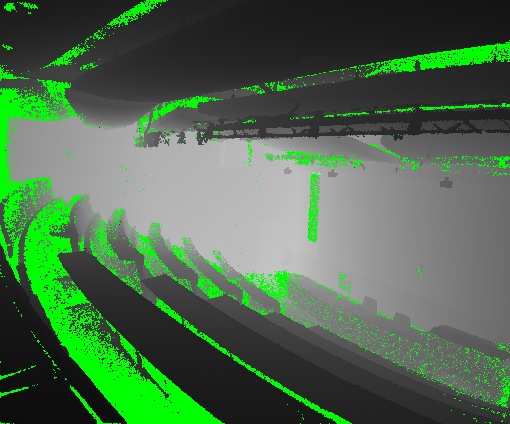}} \\
		\scalebox{-1}[1]{\includegraphics[width=0.325\columnwidth]{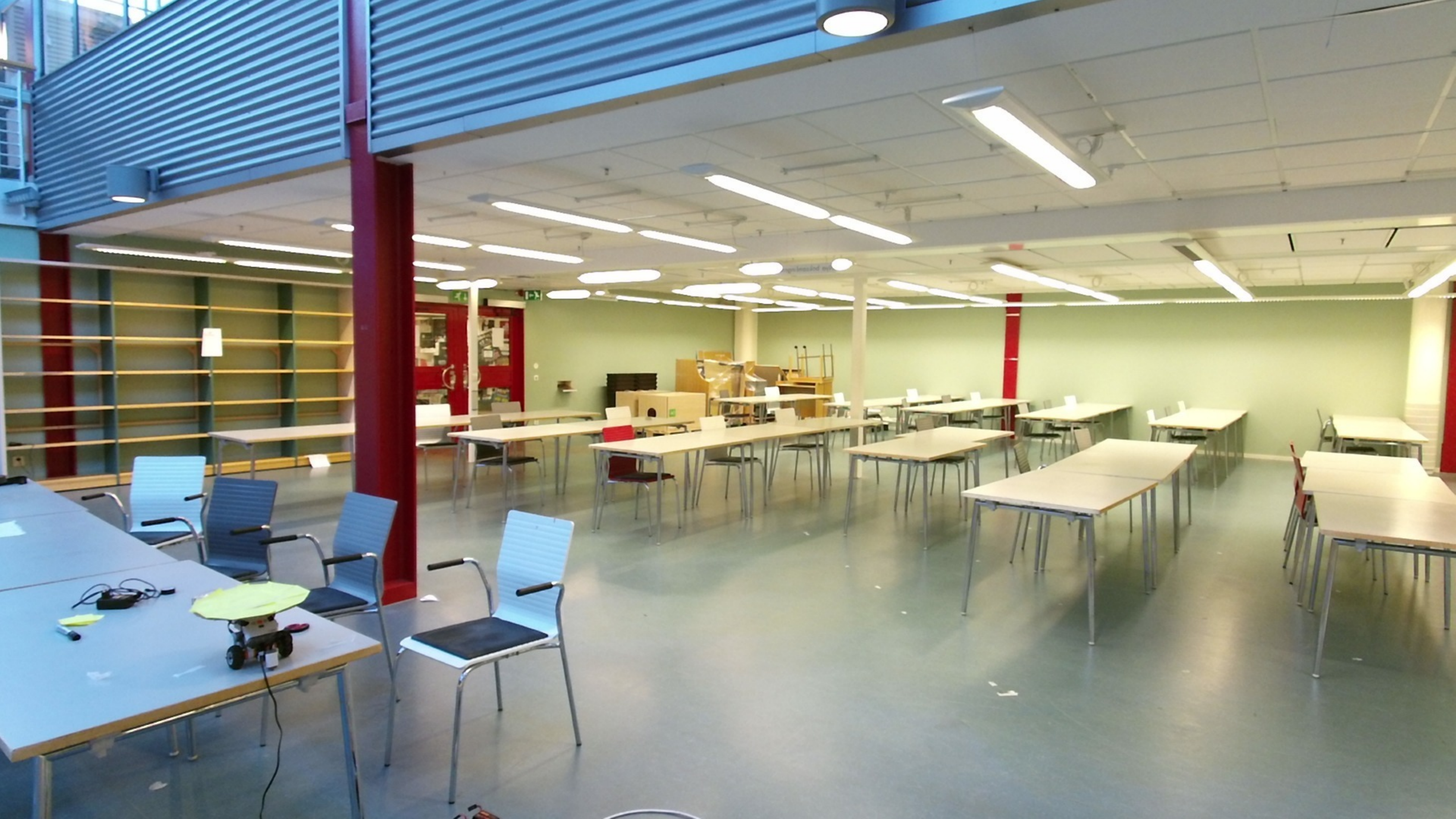}} &
		\includegraphics[width=0.325\columnwidth]{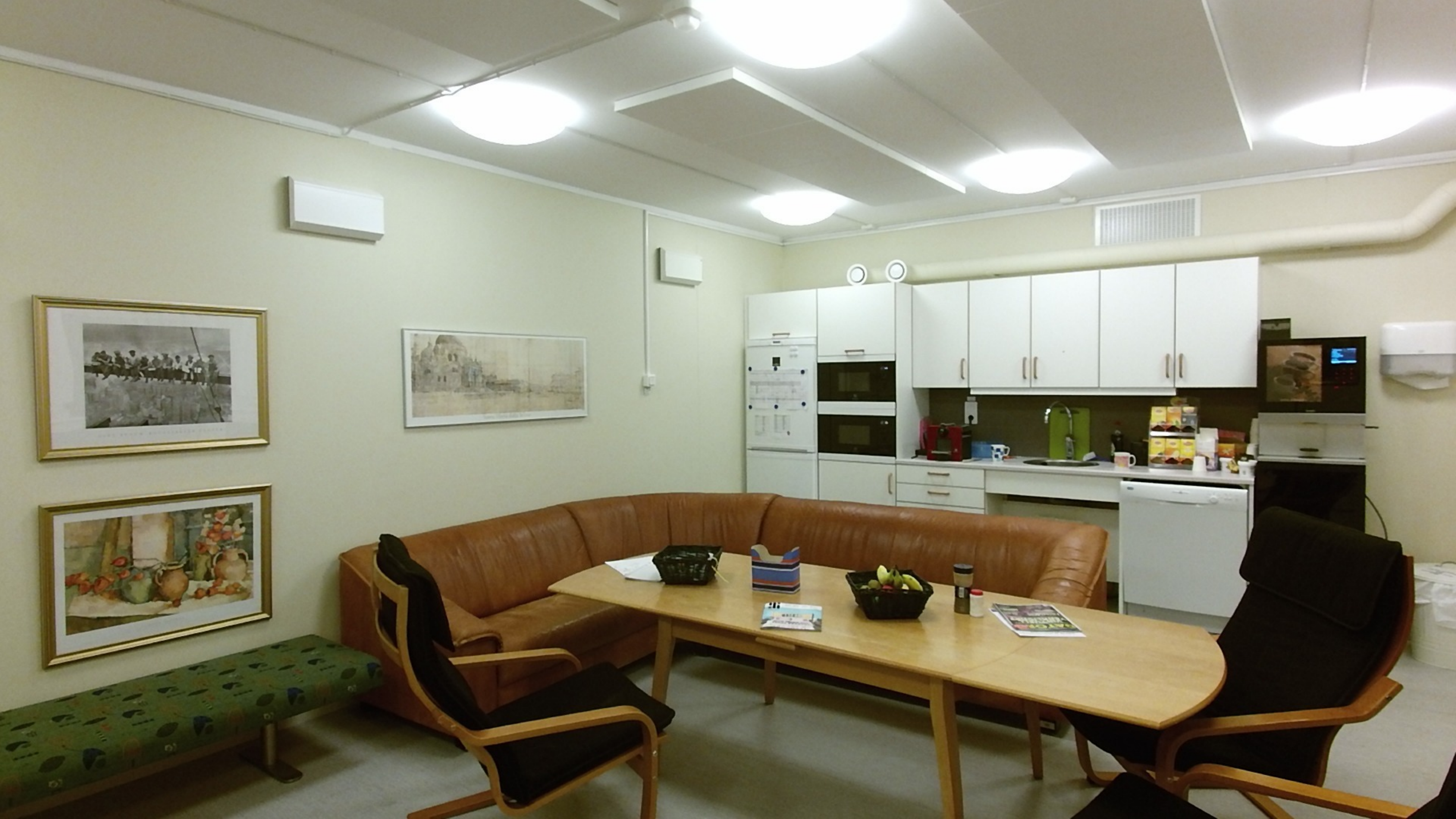} &
		
		\includegraphics[width=0.325\columnwidth]{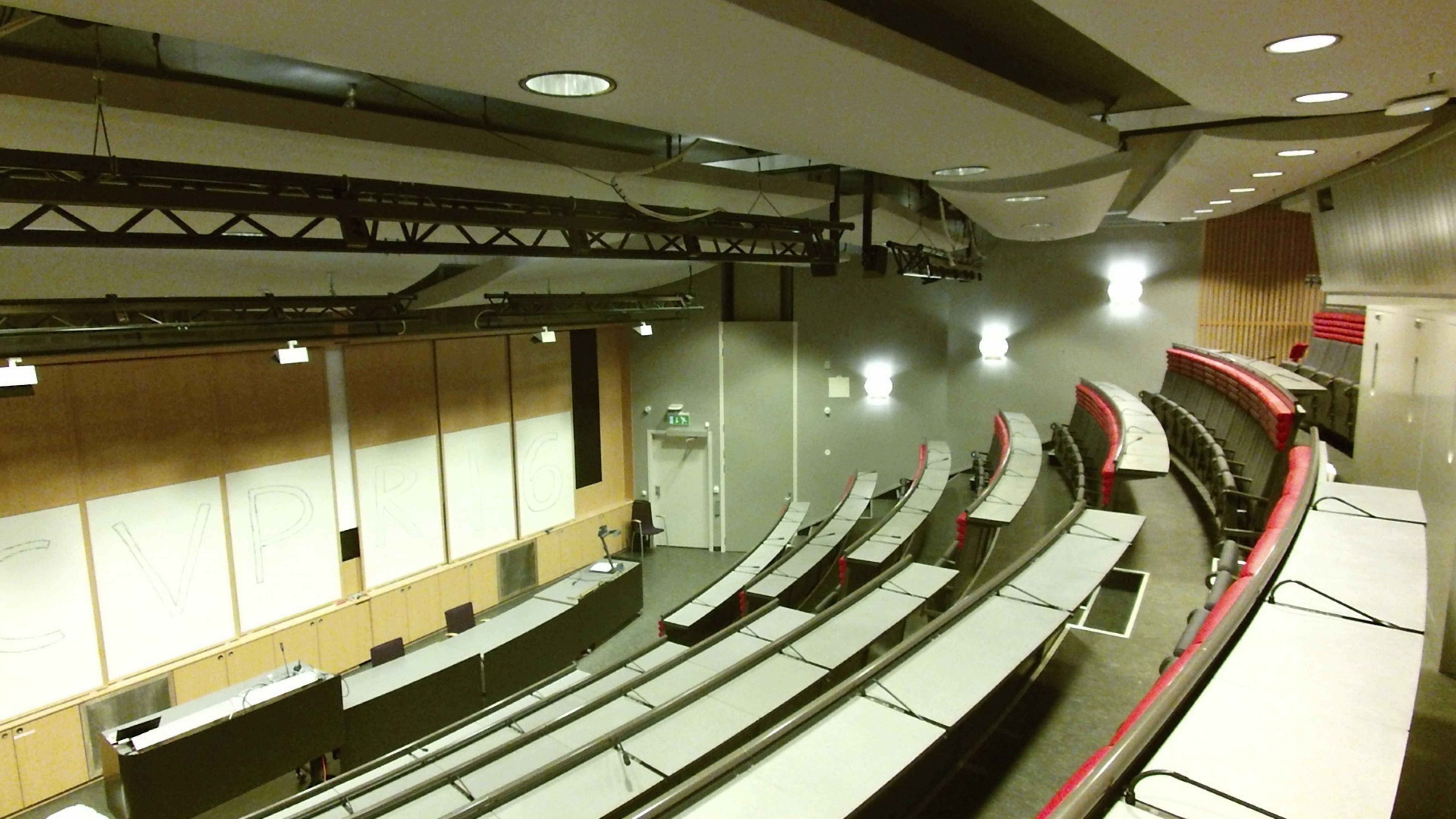} \\
	\end{tabular}
	\caption{Unwrapping ground truth for the three datasets.
		Top row: ground truth depth maps. Green pixels are
		suppressed, and not used in the evaluation. Bottom row:
		corresponding images from the RGB camera.}
	\label{fig:ground_truth}
\end{figure}

\subsection{Ground Truth for Unwrapping}
\label{sec:ground_truth}
We construct our own ground truth data, which is used for quantitatively evaluating the correctness of the phase unwrappings. The accuracy of
the ground truth must be good enough to tell a correct unwrapping from
an incorrect one. As we have 30 unwrapping candidates in an 18.75m
range, the distance between the candidates is on average 60cm. To
ensure that no incorrect unwrappings are accidentally counted as
inliers, we require an accuracy of at least half the candidate distance,
i.e.~better than 30cm.

The required accuracy can easily be met using the Kinect sensor itself.
By fusing many frames from the same camera pose, we can reduce the
amount of unwrapping errors, and also increase the accuracy in
correctly unwrapped measurements. For a given scene we place the camera at different locations corresponding to a spatial $3\times 3$ grid. By fusing data
from these poses we can detect and suppress multipath
responses, which vary with camera position. Further details on the dataset generation can be found in the supplemental materials \cite{jaremo-lawin16a}.

\subsection{Datasets}
\label{sec:datasets}
We have used the procedure in section \ref{sec:ground_truth} to
collect three datasets with ground truth depth, shown in figure
\ref{fig:ground_truth}. The {\bf kitchen} dataset has a maximal depth
of $6.7$m, and is used to test the Kinect v2 under the intended usage
with an $8$m depth limit. The {\bf lecture} dataset has a maximal
depth of $14.6$m and is used to evaluate methods without imposing the
$8$m limit. The {\bf library} dataset is used for parameter tuning,
and has a maximal range of $17.0$m. For each dataset, we have
additionally logged $25$ raw-data frames from the central camera pose,
using a data logger in Linux, and another $25$ output frames using the
{\it Microsoft SDK v2} API in Windows.

\subsection{Comparison of noise propagation models}
\label{sec:noise_models}
\begin{figure}[!t]
	\begin{tabular}{@{}c@{\;}c@{\;}c@{}}
		\includegraphics[width=0.32\columnwidth, trim= 40 -18 40 0, clip]{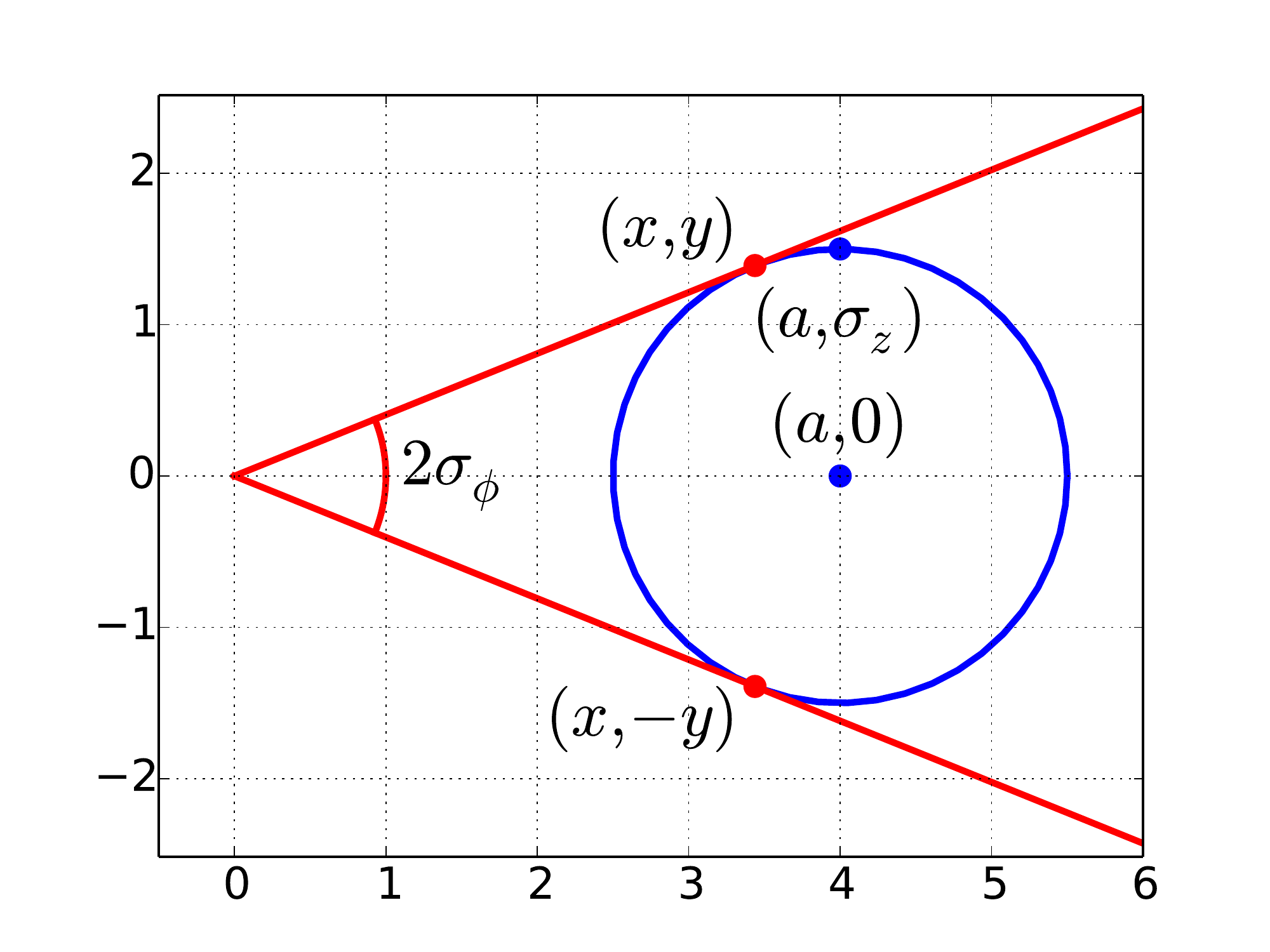} &
		\includegraphics[height=3.2cm]{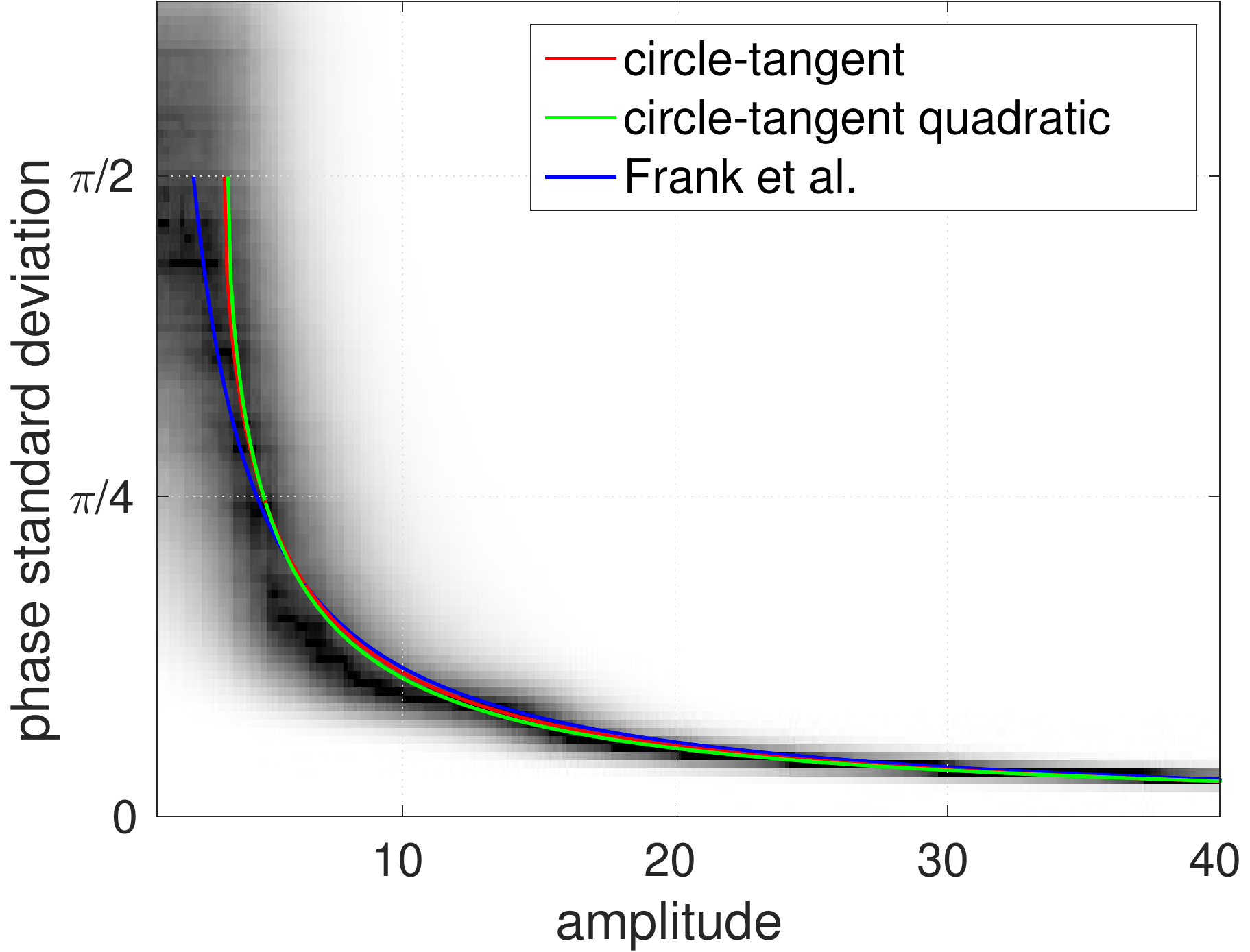} &
		\includegraphics[height=3.2cm, trim=60 0 0 0, clip]{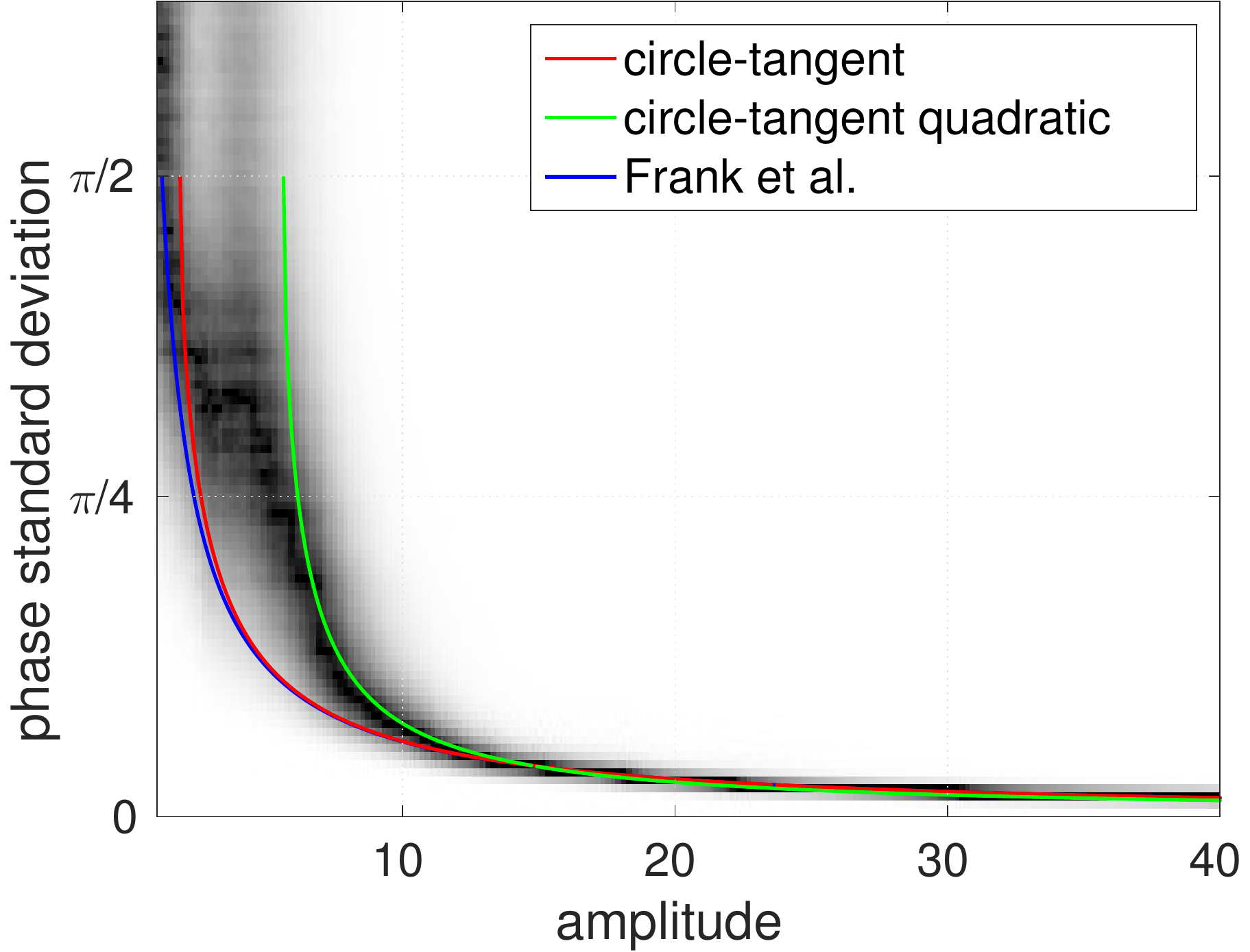}\\
		(a) &  \hspace{3mm} (b) & \hspace{-1mm}(c)
	\end{tabular}
		\caption{(a): Geometrical illustration of the circle-tangents. (b): Predictions from raw phase overlaid on
                  empirical distribution. (c): Predictions from bilateral-filtered phase.}
		\label{fig:propagation_models}

\end{figure}

The tuning dataset {\bf library} was used to estimate the standard
deviations $\sigma_{\phi}$ of the individual phase measurements over
40 frames. The model parameters in \eqref{eq:sigma_frank}, \eqref{eq:sigma_raw} and
\eqref{eq:sigma_bilateral} were found by minimizing the residuals of the corresponding inverted expressions using non-linear least squares over all amplitude measurements $a$. The inversion of the expressions reduced bias effects due to large residuals for small amplitudes. 

This procedure was
performed for ${\bf z}$ with and without bilateral filtering (as
implemented in {\it libfreenect2}). Figure \ref{fig:propagation_models} ((b) and (c)) shows the resulting predictions overlaid on the empirical distributions of the relation between the amplitude and the phase standard deviation. We see that the models proposed in \eqref{eq:sigma_raw} and \eqref{eq:sigma_bilateral} have a slightly better fit to the empirical distribution than \cite{frank09} on raw phase
measurements. However, for bilateral filtered ${\bf z}$, the quadratic
model suggested in expression \eqref{eq:sigma_bilateral} has the best
fit. As bilateral filtering improves the final performance this is the
model used in our method.

\subsection{Outlier Rejection}

\textbf{\textit{libfreenect2}}: outlier rejection is performed at several
steps, each with one or several tuned thresholds:
\begin{itemize}
	\item Pixels where any of the amplitudes is below a threshold are suppressed.
	\item Pixels where the pseudo-distances differ in magnitude are
	suppressed. The purpose of this is similar to \eqref{eq:unwrapping_likelihood},
	but an expression based on the cross-product of the pseudo phases
	with a reference relation is used.
	\item Pixels with a large depth, or amplitude variance in their
	$3\times 3$ neighbourhood are suppressed.
	\item Pixels that deviate from their neighbours are suppressed.
	\item Pixels on edges in the voltage images are suppressed.
\end{itemize}
{\bf Proposed method}: a single threshold is applied on the KDE-based confidence measure in \eqref{eq:confidence}.

\subsection{Parameter settings}
The proposed method introduces the following parameters that needs to be set:
\begin{itemize}
	\item the scaling $s_1$ in \eqref{eq:unwrapping_likelihood}
	\item the scaling $s_2$ in \eqref{eq:phase_likelihood}.
	\item the kernel scale $h$ in \eqref{eq:the_kernel}.
	\item the spatial support $r$. (the Gaussian in \eqref{eq:weights} has a spatial support of $(2r+1)\times(2r+1)$ and $\sigma=r/2$.)
	\item the number of hypotheses $|\mathcal{I}|$.
\end{itemize}
The method is not sensitive to the selection of $s_1$, $s_2$ and $h$,
and thus the same setting is used for all experiments. Unless otherwise stated, the parameters $r=5$ and $|\mathcal{I}|=2$ are used. The effects of these parameters are discussed further in the supplemental material \cite{jaremo-lawin16a}.

\subsection{Coverage Experiments}
\begin{figure}[!t]
	\begin{tabular}{@{}c@{\;}c@{\;}c@{}} 
		\hbox{\rule{3ex}{0pt}}{\bf kitchen} & {\bf kitchen} (depth limited) & {\bf lecture} \\
		\includegraphics[height=0.27\columnwidth]{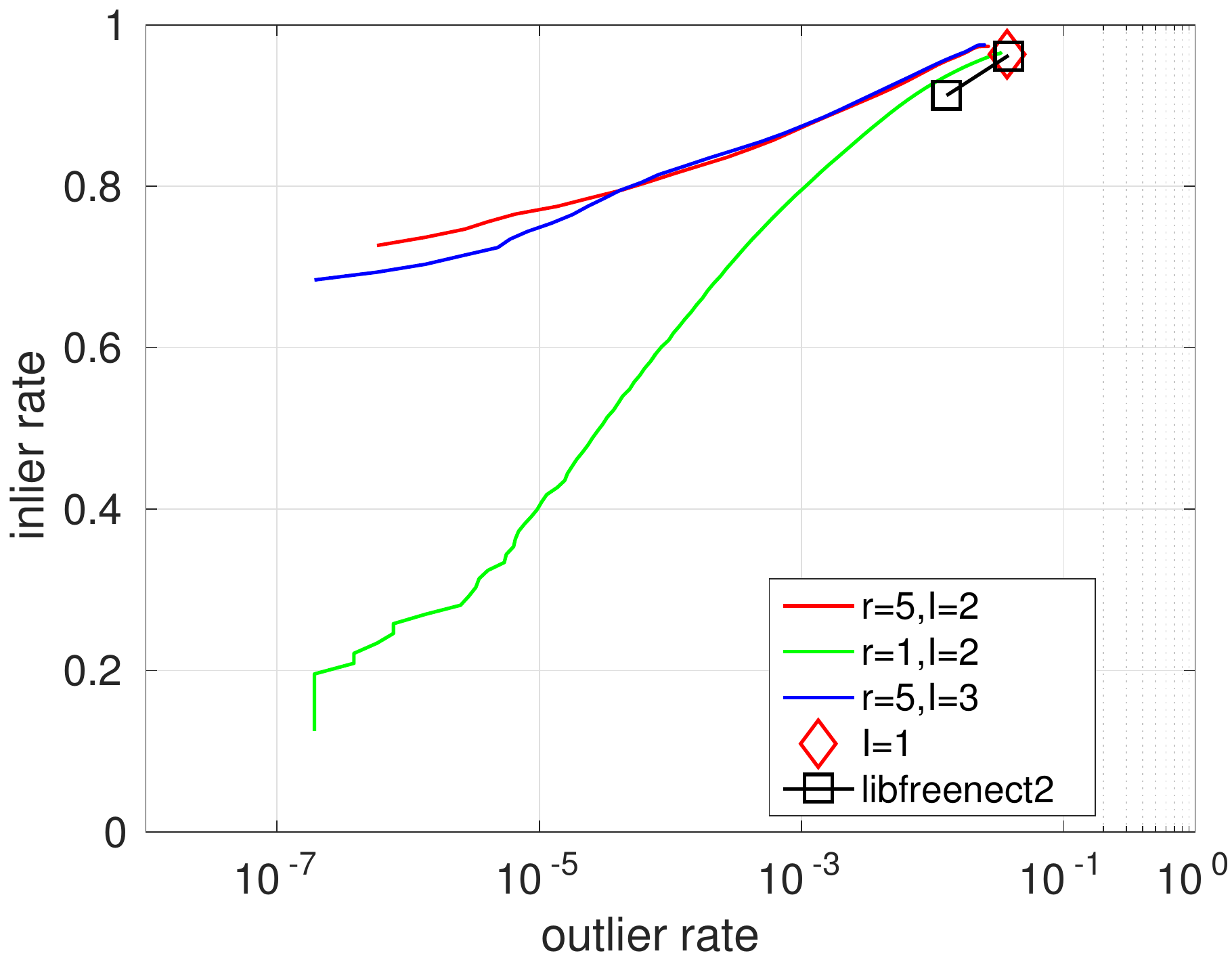} &
		\includegraphics[height=0.27\columnwidth,trim=60 0 0 0, clip]{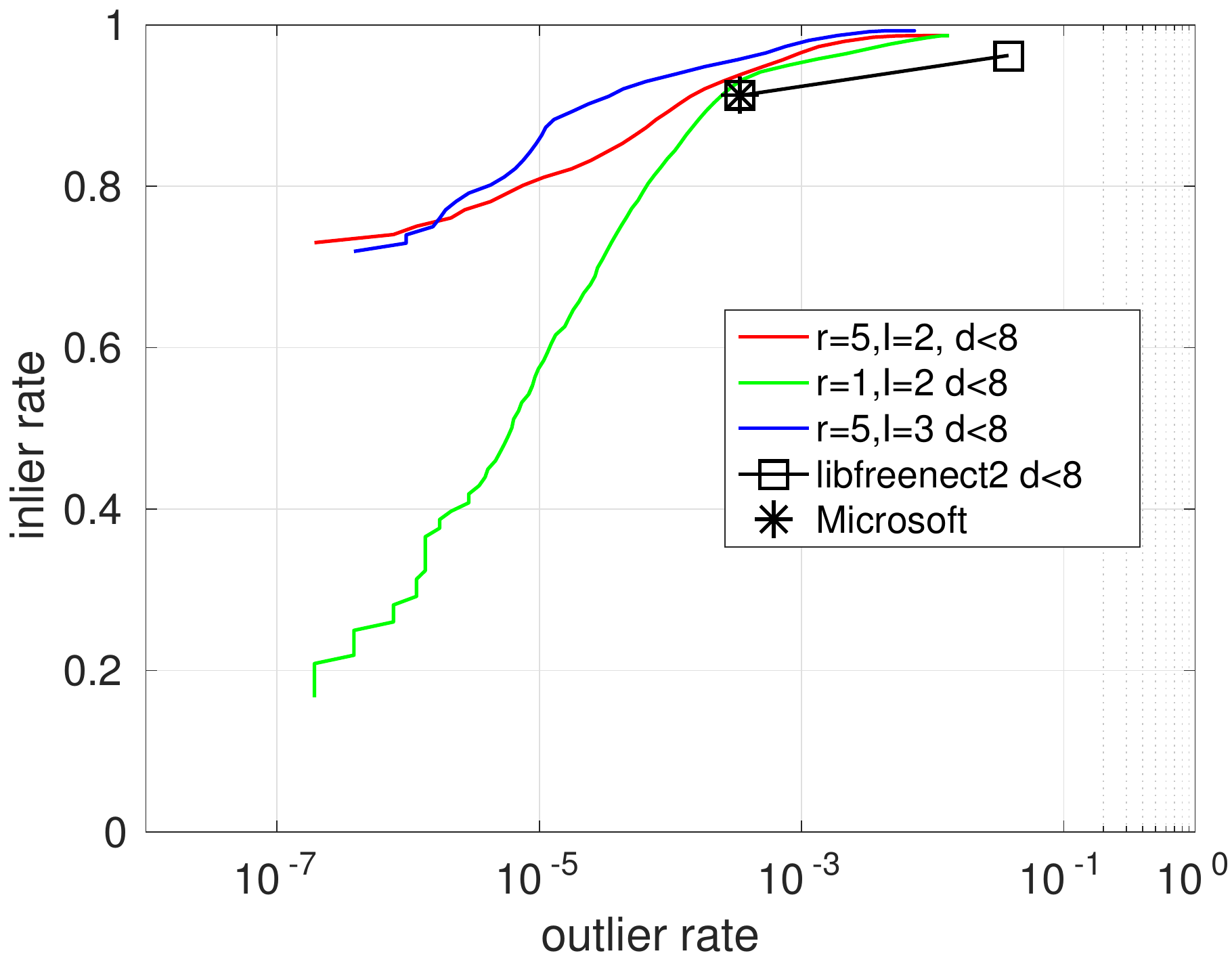}&
		\includegraphics[height=0.27\columnwidth,trim=60 0 0 0, clip]{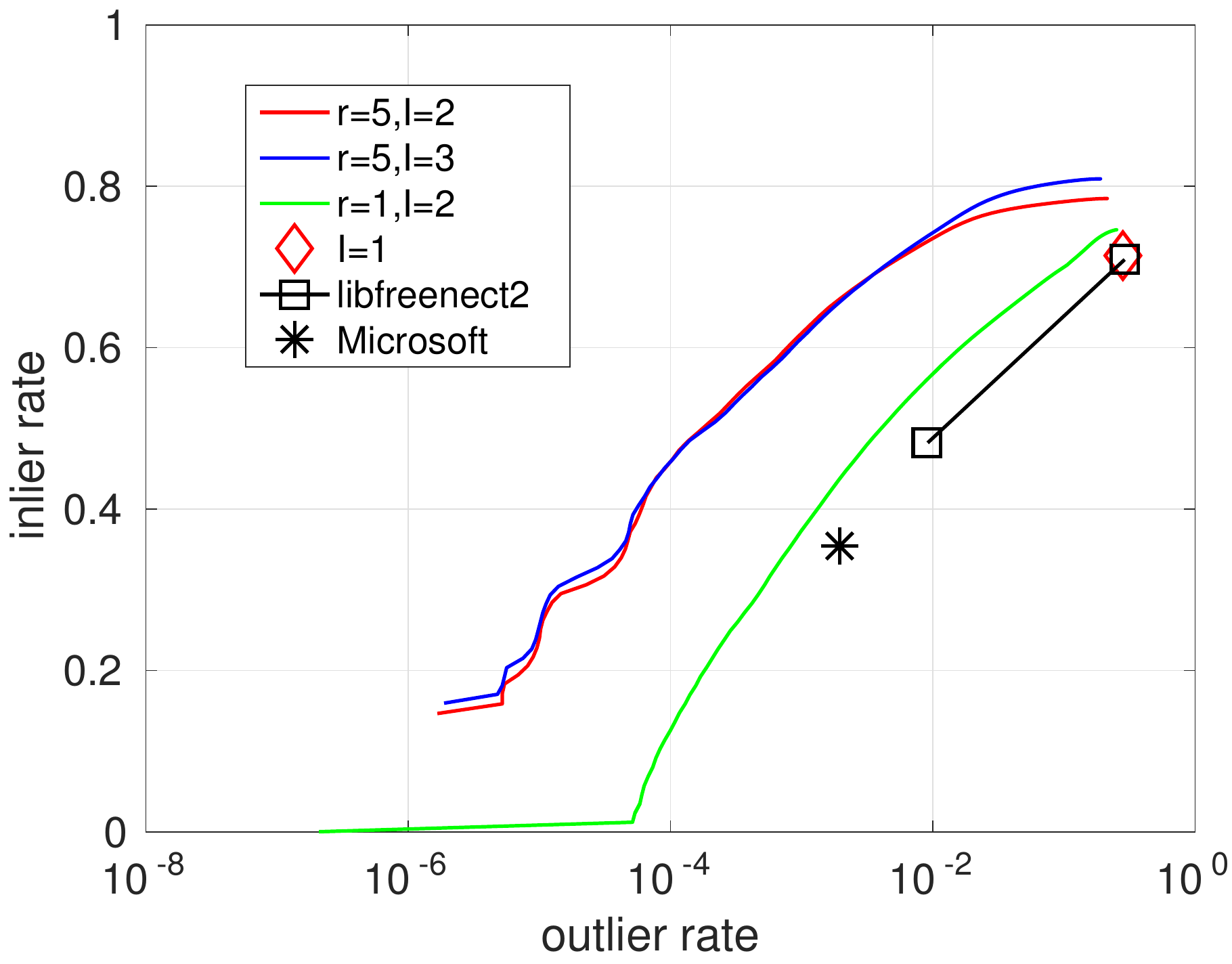}
	\end{tabular}
	\caption{Inlier and outlier rate plots. Each point or curve is the average over 25 frames.}
	\label{fig:results}
\end{figure}
We have used the unwrapping datasets described in section
\ref{sec:datasets} to compare the methods in terms of inliers
(correctly unwrapped points), and outliers (incorrectly unwrapped
points). A point is counted as an inlier when a method outputs a depth
estimate which is closer than 30cm to the ground truth, and an outlier
otherwise. These counts are then divided by the number of valid points
in the ground truth to obtain inlier and outlier rates.

Figure \ref{fig:results} shows plots of inlier rate against
outlier rate for our method, for the full range of thresholds on the
output confidence in \eqref{eq:confidence}. As a reference, we
also plot the output from {\it Microsoft} and {\it libfreenect2},
as well as {\it libfreenect2} without the outlier threshold, and {\it
	libfreenect2} where the hypothesis selection is done by minimising
\eqref{eq:knm_cost}, instead of using the CRT approach in section
\ref{sec:greedy} (labelled $I=1$ in the legend). As can be seen in
figure \ref{fig:results} middle, the performance of {\it libfreenect2}
and {\it Microsoft} are similar on short range
scenes with a depth limit (this is expected, as the {\it libfreenect2} source mentions it being based on disassembly of the Microsoft SDK). 

As can be seen, the proposed method consistently has a higher inlier rate at the same outlier rate, when compared to {\it libfreenect2} with the same spatial support, i.e.~$r=1$. When the spatial support size is increased, the improvement is more pronounced.

Performance for scenes with larger depth is exemplified with the {\bf lecture} dataset. With the depth limit removed, we get significantly more valid measurements at the same outlier rate. The {\it Microsoft} method has a hard limit of $8$m and cannot really compete on this dataset; it only reaches about
$35\%$ inlier rate. The {\it libfreenect2} method without the depth limit reaches $48\%$ inliers, at a $1\%$ outlier rate. At the $1\%$
outlier rate, the proposed method has a $73\%$ inlier rate, which is
an relative improvement of $52\%$ over {\it libfreenect2}.

The performance is improved slightly for $|\mathcal{I}|=3$ compared
with $|\mathcal{I}|=2$. While still having frame rates over 30 fps for
a spatial support of $r = 5$, we consider the costs to outweigh the
small improvement, and thus favour the setting of $|\mathcal{I}|=2$. 

\subsection{Kinect Fusion}

\begin{figure}[!t]

	\scalebox{-1}[1]{\framebox{\includegraphics[width=0.475\columnwidth,trim=0 0 40 0]{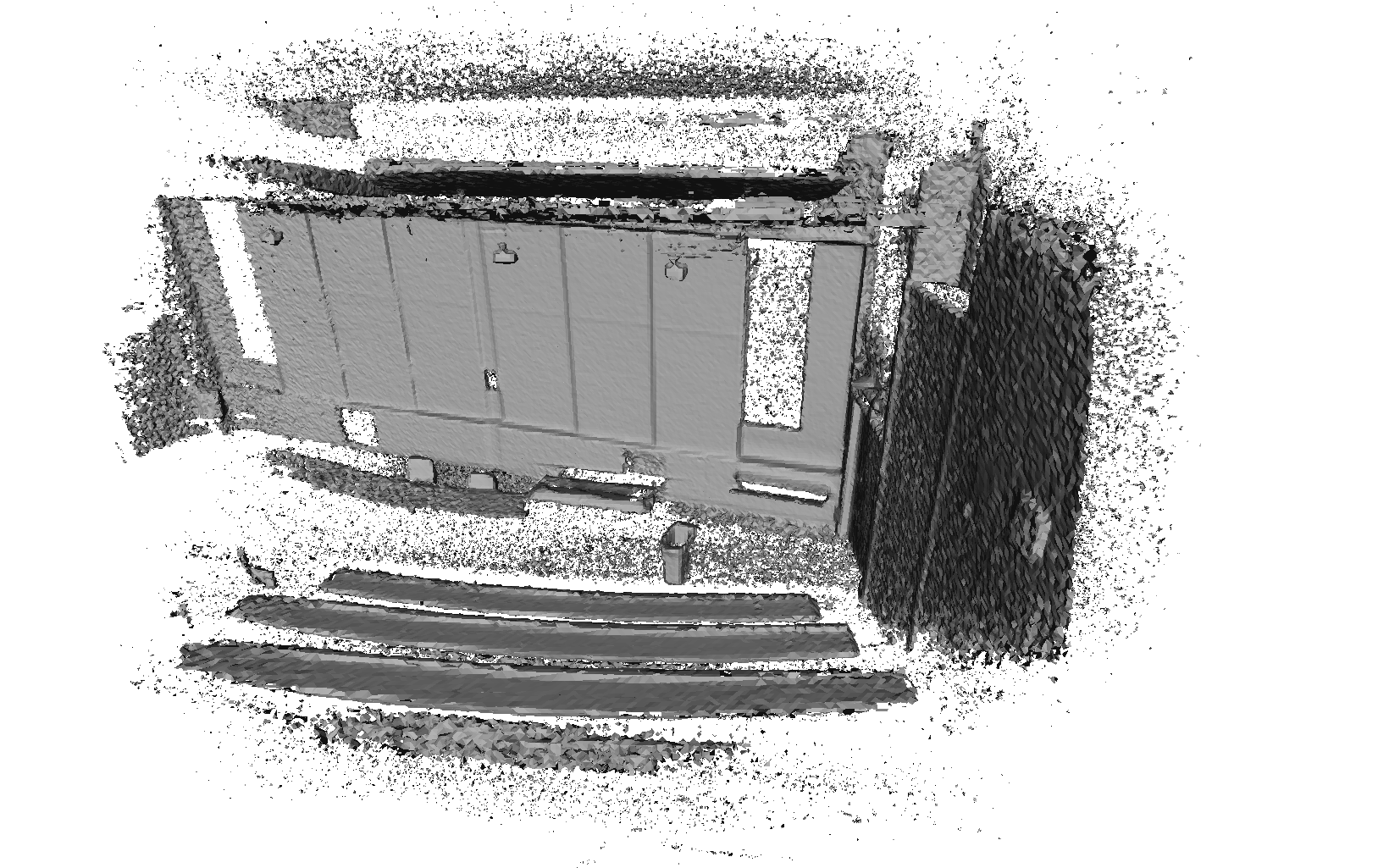}}}
	\scalebox{-1}[1]{\framebox{\includegraphics[width=0.475\columnwidth,trim=0 0 40 0]{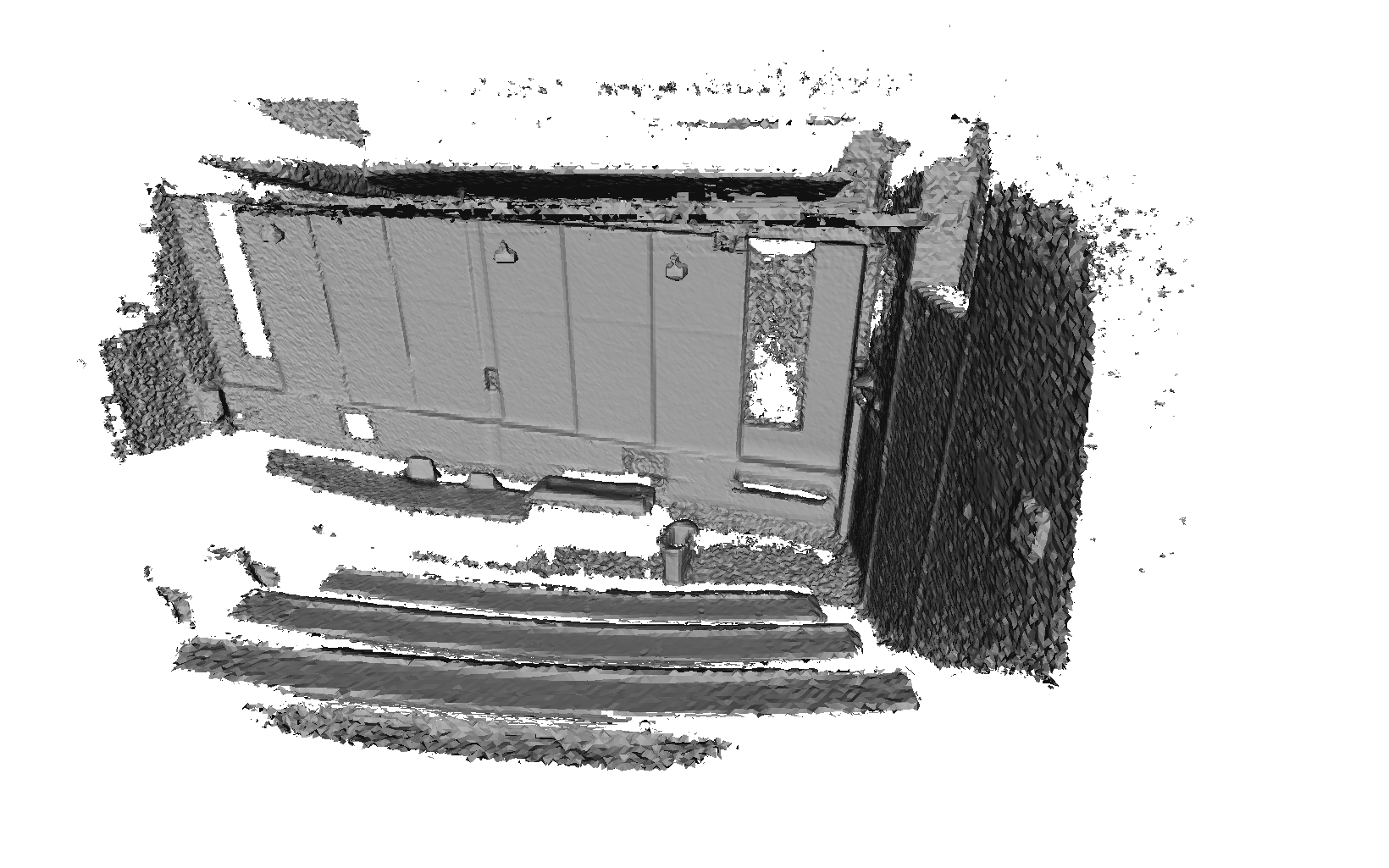}}}

	\caption{Meshes of {\bf lecture} scene from KinFu. Left: unwrapped with
		{\it libfreenect2}. Right: unwrapped with the proposed method.}
	\label{fig:meshes}
\end{figure}

We have implemented a data-logger that saves all output from the
Kinect v2 to a file for later playback. This allows us to feed the
Kinect Fusion implementation KinFu in the {\it Point Cloud
	Library} \cite{Rusu_ICRA2011_PCL} with Kinect v2 output unwrapped
with both {\it libfreenect2} and the proposed method. Figure
\ref{fig:meshes} shows two meshes obtained in this way.
As can be seen Kinect Fusion benefits from the proposed approach by
generating models with fewer outlier points, and consistently more
complete scene details. See \cite{jaremo-lawin16a} for more examples.

\section{Concluding Remarks}
\label{sec:conclusions}
This paper introduces a new multi-frequency phase unwrapping
method based on kernel density estimation of phase hypotheses in a spatial neighbourhood. We also derive a new closed-form expression for prediction of
phase noise and show how to utilize it as a measure of confidence for the measurements.

Our method was implemented and tested extensively on
the Kinect v2 time-of-flight depth sensor. Compared to the previous
methods in {\it libfreenect2} and {\it Microsoft Kinect SDK v2} it
consistently produces more valid measurements when using the default
depth limit of $8$m, while maintaining real-time performance. In
large-depth environments, without the depth limit, the gains are
however much larger, and the number of valid measurements increases by
$52\%$ at the same outlier rate.

As we have shown, the proposed method allows better 3D scanning of
large scenes, as the full $18.75$m depth range can be used. This
is of interest for mapping and robotic navigation, where seeing
further allows better planning. As the method is generic, future work includes applying it to other multi-frequency problems such as Doppler radar \cite{trunk93} and fringing \cite{wang2009,gorthi10}. 
\subsubsection{Acknowledgements}
This work has been supported by the Swedish Research Council in
projects 2014-6227 (EMC2) and 2014-5928 (LCMM) and the EU's Horizon 2020 Programme grant No 644839 (CENTAURO).

\bibliographystyle{splncs}
\bibliography{main}
\end{document}